\documentclass[conference]{IEEEtran}
\IEEEoverridecommandlockouts
\usepackage{url}
\usepackage{cite}
\usepackage{amsmath,amssymb,amsfonts}
\usepackage[vlined,ruled,linesnumbered]{algorithm2e}
\SetAlgoCaptionSeparator{.}

\usepackage{graphicx}
\usepackage{textcomp}
\usepackage{xcolor}
\usepackage{microtype}
\usepackage{booktabs}

\def\BibTeX{{\rm B\kern-.05em{\sc i\kern-.025em b}\kern-.08em
    T\kern-.1667em\lower.7ex\hbox{E}\kern-.125emX}}
\begin{document}

\setlength{\fboxsep}{1pt}

\title{Online Diversity Control in Symbolic Regression via a Fast Hash-based Tree Similarity Measure
\\
\thanks{The authors gratefully acknowledge support by the Christian Doppler Research Association and the Federal Ministry of Digital and Economic Affairs within the \emph{Josef Ressel Centre for Symbolic Regression}}}

\author{\IEEEauthorblockN{Bogdan Burlacu$^{1,2}$, Michael Affenzeller{$^2$}, Gabriel Kronberger$^{1,2}$, Michael Kommenda$^{1,2}$}
\IEEEauthorblockA{
$^1$\textit{Josef Ressel Centre for Symbolic Regression}\\ 
$^2$\textit{Heuristic and Evolutionary Algorithms Laboratory} \\
\textit{University of Applied Sciences Upper Austria}\\
Softwarepark 11, 4232 Hagenberg, Austria}
bogdan.burlacu@fh-hagenberg.at\\
michael.affenzeller@fh-hagenberg.at\\
gabriel.kronberger@fh-hagenberg.at\\
michael.kommenda@fh-hagenberg.at
}

\maketitle

\begin{abstract}
    Diversity represents an important aspect of genetic programming, being directly correlated with search performance. When considered at the genotype level, diversity often requires expensive tree distance measures which have a negative impact on the algorithm's runtime performance.
    In this work we introduce a fast, hash-based tree distance measure to massively speed-up the calculation of population diversity during the algorithmic run. We combine this measure with the standard GA and the NSGA-II genetic algorithms to steer the search towards higher diversity. We validate the approach on a collection of benchmark problems for symbolic regression where our method consistently outperforms the standard GA as well as NSGA-II configurations with different secondary objectives.
\end{abstract}



\section{Introduction}
Many studies in the field of genetic programming affirm the key role of population diversity in avoiding premature convergence and improving search performance. Some of them consider diversity in terms of program behaviour (phenotypic diversity)~\cite{Rosca-entropy-drivenadaptive,matsui1999new,1144290,1299834,Ahmad:2018:GECCOcomp,10.1007/978-3-642-15871-1_48,10.1007/978-3-642-29139-5_10} while others consider diversity in terms of program structure (genotypic diversity)~\cite{oreilly:1997:dnGPugo2,Wiese:1998:KRS:330560.330837,McPhee:1999:AGD:2934046.2934071,Ekart:2000:MGP:646808.703952,Burks:2016:GPEM}. 

Vanneschi et al.~\cite{Vanneschi2014} observe that calculating the semantics of a program is a side effect of fitness calculation, available at no extra computational cost, and argue that semantic diversity is more helpful than genetic diversity. This opinion is shared by Burke et al.~\cite{1266373} who observe that ``when a many-to-one relationship exists between the genotype and phenotype encoding, measures which are based on genotype uniqueness will probably not be as useful as those which capture phenotype uniqueness''. 
%

We argue that redundancy in genotype space, described as a many-to-one relationship between genotypes and phenotypes represents a prerequisite for evolvability (defined as the ability of random variations to sometimes produce improvement). 

Ebner et al.~\cite{Ebner:2001:NNI:500120.500125} show that redundancy in this context is important as it increases the accessibility between phenotypes via neutral networks in genotype space. They also find that redundancy correlates with higher genotypic diversity, thus facilitating adaptation as more diverse genotypes represent the basis for diverse behaviours. They intuitively describe this relationship: ``the smoother the landscape the easier it is to climb on top of the landscape.''

Kitano et al.~\cite{kitano2004biological} show that genotypic robustness improves evolvability when genotypes are sufficiently diverse, since robustness implies redundancy (neutral networks) in genotype space, while evolvability depends on the ability to walk across these networks towards points with higher adaptive potential. 

Genetic robustness can evolve by two main mechanisms: buffering and modularity, both conferring phenotypes a selective advantage. Buffering, through the accumulation of hidden genetic variation leads to a size increase in the genotype and the occurrence of bloat, but at the same time protects phenotypes against deleterious genotype changes and acts as an evolutionary capacitance~\cite{hayden2011cryptic}. Modularity is another way of maintaining phenotypic function against perturbation, as genotypes organised in a network of autonomous modules are less likely to change their phenotypic expression when perturbed. 

Despite decreasing phenotypic variability, robustness is a prerequisite for evolvability. Hu and Banzhaf show that robust genotypes play a crucial role in the evolutionary process as they are visited more often and can guide the search to their adjacent phenotypes~\cite{Hu:2016:GPTP,Hu:2016:GECCO}.

More recent work~\cite{Burks2018} suggests that it may be worthwhile pursuing hybrid approaches for simultaneously preserving both structural and behavioral diversity. 

We introduce a hybrid approach where an inherently structural tree distance measure is made semantically-aware by including the numerical coefficients of leaf nodes into the distance computation. 
The main idea is to assign each solution candidate a diversity score calculated as the average tree distance from the rest of the population. 

As the distance matrix for the entire population needs to be calculated in each generation, fast calculation of tree distances is key issue to this approach. Our algorithm relies on the efficient computation of hash trees for each individual in the population, and supports the computation of both structural and hybrid distances. Section~\ref{sec:methodology} describes the algorithm in detail. 

We test our approach on a collection of symbolic regression benchmark problems, using the standard Genetic Algorithm (GA) where the diversity score is added as a penalty term to fitness during selection, and the Nondominated Sorting Genetic Algorithm (NSGA-II)~\cite{996017} where we incorporate our diversity score as a secondary objective. Section~\ref{sec:results} presents the obtained results, while Section~\ref{sec:conclusion} discusses the merits as well as further applications of the approach.

We note that for clarity, the term Genetic Algorithm is used in the remainder of this contribution as a synonym for GP, as we treat tree-based GP as another problem representation of a GA.

\section{Methodology}\label{sec:methodology}
We integrate our approach within the open source optimization framework \mbox{HeuristicLab}~\cite{wagner2014} which already provides implementations of the GA and NSGA-II algorithms. The source code for the described algorithms is available online\footnote{\url{https://dev.heuristiclab.com/trac.fcgi/browser/trunk/HeuristicLab.Problems.DataAnalysis.Symbolic/3.4/Hashing}}.

We define the distance between two trees as the ratio between the number of common nodes and the total size of the two trees:
\begin{align}
    S(T_1,T_2) &= \dfrac{2 \cdot |M|}{|T_1| + |T_2|}~(\mathrm{tree~similarity})\label{eqn:tree-similarity}\\
    D(T_1,T_2) &= 1 - S(T_1, T_2)~(\mathrm{tree~distance})\label{eqn:tree-distance}
\end{align} 
where $M$ represents a mapping of isomorphic subtrees from $T_1$ to $T_2$. The mapping is computed by transforming each tree into a sequence of integer hash values and then identifying pairs of subtrees with the same hash in both $T_1$ and $T_2$.


\subsection{Tree Hashing}\label{subsec:tree-hashing}

The proposed tree hashing algorithm shares some common aspects with Merkle trees~\cite{10.1007/3-540-48184-2_32} -- an encryption scheme where every leaf node is labelled with the hash of a data block and every non-leaf node is labelled with the hash of the labels of its child nodes. In our approach, each non-leaf tree node is assigned an initial hash value which is then aggregated with the hash values of its child nodes. If the node represents a commutative operation, its child nodes are sorted in order to ensure consistent hashing over different argument orders.
Algorithm~\ref{alg:tree-hashing} provides a high-level overview of the procedure. The following notations are used:
\begin{itemize}
    \item \texttt{Postorder}($T$) - $T$'s nodes visited in postorder
    \item \texttt{Hash}$(input)$ - hash function used by the algorithm
    \item $H(n)$ - the hash value of node $n$ 
\end{itemize}

The traversal of $T$ in postorder ensures that its nodes are sorted and hashed in a single bottom-up pass. The child order for commutative nodes is established by simple precedence rules (internal node before leaf, constant before variable node, and so on). Nodes of the same type are ordered based on their hash value. 

Implementation-wise, the nodes of $T$ in postorder are stored as an array which is then iterated over from left to right. This representation has the advantage of simplicity as all tree operations can be expressed using basic arithmetic between array indices. For example, am internal node $n$ at position $i$ in the array will find its first child at index $j=i-1$. The index of the next child is obtained by subtracting the size of the first child subtree from $j$, and so on. 

Sorting child nodes in the array representation (Algorithm~\ref{alg:tree-hashing}, line~\ref{alg:tree-hashing-sort}) is equivalent to putting the corresponding subarrays in the correct order. In the simple case (when all child nodes are leaves) a single sort operation is necessary. For non-leaf nodes the subarrays must be reordered which involves two additional copy operations using an auxiliary buffer. 
After sorting child node hash values are aggregated with the current node's hash value (Algorithm~\ref{alg:tree-hashing}, line~\ref{alg:tree-hashing-compose-hash}). The \texttt{Hash} aggregation function can be either a general-purpose or a cryptographic-strength method. Our current implementation uses a popular hash function known as \texttt{DJB}\footnote{\url{http://www.partow.net/programming/hashfunctions/#DJBHashFunction}} and illustrated in Algorithm~\ref{alg:hash-function}.

\begin{algorithm}
    \small
    \caption{\sc{\footnotesize Tree hash algorithm}}\label{alg:tree-hashing}
    \SetKwInOut{Input}{input}\SetKwInOut{Output}{output}
    \SetKwFunction{Postorder}{Postorder}
    \SetKwFunction{Children}{Children}
    \SetKwFunction{Root}{Root}
    \SetKwFunction{Hash}{Hash}
    \SetKw{Continue}{continue}
    \SetKw{Return}{return}
    \Input{An expression tree $T$}
    \Output{The corresponding sequence of hash values}
    \BlankLine
    hashes $\gets$ empty list of hash values\;
    \ForEach{node $n$ in \Postorder(T)}
    {
        $H(n) \gets$ an initial hash value\;
        \If{$n$ is an internal node}
        {
            \If{$n$ is commutative}{Sort the child nodes of $n$\;\label{alg:tree-hashing-sort}}
            child hashes $\gets$ hash values of $n$'s children\;
            $H(n) \gets$ \Hash{\textup{child hashes}, $H(n)$}\;\label{alg:tree-hashing-compose-hash}
        }
        hashes.append($H(n)$)\;
    }
    \Return hashes\;
\end{algorithm}


Finally, Algorithm~\ref{alg:tree-hashing} returns a list of hash values corresponding to the tree nodes visited in postorder. 

\begin{algorithm}
    \small
    \caption{\sc{\footnotesize DJB hash function}}\label{alg:hash-function}
    \SetKwInOut{Input}{input}\SetKwInOut{Output}{output}
    \SetKw{Return}{return}
    \SetKw{ulong}{unsigned integer}
    \Input{A sequence of bytes}
    \Output{An aggregated hash value}
    \BlankLine
    \ulong $hash \gets 5381$\;
    \ForEach{input byte $b$}
    {
        $hash \gets (hash \ll 5) + hash + b$\;
    }
    \Return hash\;
\end{algorithm}

\subsection{Population Distance Matrix}\label{subsec:similarity-matrix}
Given Algorithm~\ref{alg:tree-hashing}, we can easily compute tree distance by simply comparing two sorted sequences of node hashes, as illustrated in Algorithm~\ref{alg:tree-distance}. The tree distance matrix for the whole population can then be computed in a few steps:

\begin{enumerate}
    \item Hash all tree individuals using Algorithm~\ref{alg:tree-hashing}.
    \item Sort all of the resulting hash arrays.
    \item Compute pairwise distances using Algorithm~\ref{alg:tree-distance}.
\end{enumerate}

The efficiency of the method comes from the fact that pairwise distances between trees are computed in linear time since Algorithm~\ref{alg:tree-distance} runs in $O(\min(|H_1|, |H_2|))$ after sorting.

The semantics of the trees can be (indirectly) taken into consideration by including the numerical coefficients of leaf nodes in the computation of the tree node hash values. The two types of leaf nodes, \emph{constant} and \emph{variable} are characterized by a numerical value and a weighting factor, respectively.

This leads to two different hashing behaviours:
\begin{itemize}
    \item Hybrid hashing: node labels as well as coefficients of numeric leaf nodes are hashed together 
    \item Structural hashing: considers only the structure of the tree by taking only node labels into account for hashing 
\end{itemize}

\begin{algorithm}
    \small
    \caption{\sc{\footnotesize Merge-count hash values}}\label{alg:tree-distance}
    \SetKwInOut{Input}{input}\SetKwInOut{Output}{output}
    \SetKw{And}{and}
    \SetKw{Return}{return}
    \SetKwFunction{Sort}{Sort}
    \Input{Two sorted hash arrays $H_1$ and $H_2$}
    \Output{The calculated distance according to Equation~\ref{eqn:tree-distance}}
    \BlankLine
    $i \gets 0$, $j \gets 0$, $count \gets 0$\;
    \While{$i < |H_1|~\And~j < |H_2|$}{
        \uIf{$H_1[i]$ = $H_2[j]$}{
            $count \gets count+1$\;
            $i \gets i+1$\;
            $j \gets j+1$\;
        }
        \uElseIf{$H_1[i] < H_2[j]$}{
            $i \gets i+1$\;
        }
        \Else{
            $j \gets j+1$\;
        }
    }
    \Return $1 - \dfrac{2 \cdot count}{|H_1| + |H_2|}$
\end{algorithm}

\subsection{Correctness and Runtime Performance}
We have described so far an algorithm for processing tree individuals into linear sequences of hash values corresponding to a postorder traversal of nodes. Like any hashing scheme, its reliability in practice depends on the hash function used and its vulnerability to hash collisions.   

The hash-based tree distance validates successfully against the baseline method, producing identical results. In terms of runtime, our new method achieves a significant speed-up as shown in Table~\ref{tab:distance-performance}.  

We test the algorithm's correctness against the bottom-up tree distance by Valiente~\cite{Valiente01anefficient} that runs in time linear to the sizes of the two trees. We generate 5000 tree individuals (amounting to $\approx$12.5 million distance calculations) and calculate the average distance using both methods, then compare the resulting values as well as the running time of the two algorithms. 

We measure the performance of the hash-based tree distance both in batch-mode (where each tree is only hashed once) and in single-mode, where each distance calculation hashes the two trees anew. Incidentally, this also shows that theoretical runtime complexity does not guarantee good performance. For example, the bottom-up tree distance relies on multiple dictionary look-ups in its implementation, while we rely on a linear data structure and a fast and efficient hash-function.

In batch-mode, the hash-based tree distance represents a suitable tool for the online monitoring of average population diversity during the run of the algorithm.

\begin{table}
    \caption{Runtime performance of hash-based tree distance vs the bottom-up tree distance}\label{tab:distance-performance}
    \centering
    \begin{tabular}{lrr}
        \toprule
        Tree distance method & Elapsed time (s) & Speed-up\\
        \midrule
        Bottom-up                & $1225.751$ & $1.0x$\\
        Hash-based (single-mode) & $297.521$ & $4.1x$\\
        Hash-based (batch-mode)  & $3.677$ & $333.3x$\\
        \bottomrule
    \end{tabular}
\end{table}

\section{Experimental Results}\label{sec:results}

\subsection{Algorithm Configuration}
We test the NSGA-II algorithm with the $R^2$ correlation coefficient between predicted and target values as a primary objective and different secondary objectives. We compare it against the standard genetic algorithm, with both algorithms configured as described by Table~\ref{tab:nsga-config}. 


We introduce average tree distance as a secondary objective measuring how far each solution candidate is situated from the rest of the population. Our interest here is to guide the algorithm towards promising but less-explored regions of the search space. We provide a comparison between purely structural (genotypic) and hybrid (structural/semantic) diversity measures using the two hashing implementations described in Subsection~\ref{subsec:similarity-matrix}.

We further compare average tree distance with several other secondary objectives aimed at promoting parsimony:
\begin{itemize}
    \item The tree complexity measure by Kommenda et al.~\cite{Kommenda2016} aims to improve model simplicity and parsimony by recursively calculating a complexity score based on the symbols used by the model and their positions in the tree.
    \item The nested tree size or visitation length~\cite{eurogp07:keijzer} promotes parsimony and prefer shallow model structures over deeply nested ones.
    \item Tree length promotes parsimony and penalizes large trees.
\end{itemize}

For the standard genetic algorithm we introduce average tree distance as an additive penalty term in the fitness function. For a maximization problem this leads to a penalized fitness 
\[ 
    f'=f-s
\]
where $f$ is the $R^2$ correlation coefficient and $s$ is the average similarity. The two objectives are combined without additional weighting factors. 

\begin{table}[ht]
    \caption{Standard GA and NSGA-II configuration}\label{tab:nsga-config}
    \begin{tabular}{ll}
        \toprule
        Function set & Binary functions $(+, -, \times, \div)$\\
                     & Trigonometric functions $(\sin, \cos)$\\
                     & Exponential functions $(\exp, \log)$\\
        Terminal set & $constant$, $weight \cdot variable$\\
        Max. tree depth & 12 levels\\
        Max. tree length & 50 nodes\\
        Tree initialization & Probabilistic tree creator (PTC2)~\cite{luke2000two}\\
        Population size & 1000 individuals\\
        Max. generations & 500 generations\\
        Selection & Tournament selection group size 5\\
        Crossover probability & 100\%\\
        Crossover operator & Subtree crossover\\
        Mutation probability & 25\%\\
        Mutation operator & Change symbol, single-point,\\
                          & remove branch, replace branch\\
        Primary objective & $R^2$ correlation with the target\\
        Secondary objectives & maximize hybrid tree distance\\
                             & maximize structural tree distance\\
                             & minimize recursive complexity~\cite{Kommenda2016}\\
                             & minimize tree length\\
                             & minimize nested tree length\\
                             & minimize number of variables\\
    \end{tabular}
\end{table}

The algorithms are tested on a selection of benchmark problems as recommended by White et al.~\cite{gp-benchmarks-2013}:
\begin{itemize}
    \item Poly-10 benchmark problem~\cite{10.1007/3-540-36599-0_19}
    \item Vladislavleva benchmark problems~\cite{4632147}
    \item Pagie-1 benchmark problem~\cite{doi:10.1162/evco.1997.5.4.401}
    \item Breiman-1 benchmark problem~\cite{breiman1984classification}
    \item Friedman benchmark problems~\cite{friedman1991}
\end{itemize}

We run each algorithmic configuration for 50 repetitions on each problem and report the results in terms of median normalized mean squared error $\pm$ interquartile range.

\subsection{Benchmark Results}\label{subsec:benchmark-results}

\begin{table}
    \centering
    \caption{Breiman, Friedman, Poly-10 and Pagie-1 problems - median nmse $\pm$ iqr}\label{tab:experimental-results-various}
    \begin{tabular}{lcc}
        & \textbf{Training} & \textbf{Test}\\
        \toprule
        \textbf{Breiman-I} & &\\
        \quad GA Standard                 & $0.129 \pm 0.040$ & $0.134 \pm 0.039$\\
        \quad GA Hybrid distance          & $0.115 \pm 0.018$ & $0.120 \pm 0.021$\\ 
        \quad GA Structural distance      & $0.123 \pm 0.019$ & $0.130 \pm 0.018$\\ 
        \quad NSGA-II Hybrid distance     & \fbox{$0.110 \pm 0.012$} & \fbox{$0.117 \pm 0.013$}\\
        \quad NSGA-II Structural distance & $0.149 \pm 0.036$ & $0.154 \pm 0.038$\\
        \quad NSGA-II Tree Complexity     & $0.113 \pm 0.014$ & \fbox{$0.117 \pm 0.013$}\\
        \quad NSGA-II Tree length         & $0.122 \pm 0.017$ & $0.127 \pm 0.016$\\
        \quad NSGA-II Nested tree length  & $0.121 \pm 0.014$ & $0.126 \pm 0.014$\\
        \textbf{Friedman-I} & &\\
        \quad GA Standard                 & $0.142 \pm 0.010$ & $0.143 \pm 0.008$\\
        \quad GA Hybrid distance          & $0.139 \pm 0.003$ & $0.139 \pm 0.004$\\ 
        \quad GA Structural distance      & $0.142 \pm 0.004$ & $0.141 \pm 0.004$\\ 
        \quad NSGA-II Hybrid distance     & \fbox{$0.137 \pm 0.002$} & \fbox{$0.137 \pm 0.002$}\\
        \quad NSGA-II Structural distance & $0.150 \pm 0.015$ & $0.149 \pm 0.013$\\
        \quad NSGA-II Tree Complexity     & $0.165 \pm 0.045$ & $0.160 \pm 0.034$\\
        \quad NSGA-II Tree length         & $0.153 \pm 0.019$ & $0.150 \pm 0.015$\\
        \quad NSGA-II Nested tree length  & $0.145 \pm 0.015$ & $0.145 \pm 0.015$\\
        \textbf{Friedman-II} & &\\
        \quad GA Standard                 & $0.052 \pm 0.032$ & $0.053 \pm 0.033$\\
        \quad GA Hybrid distance          & $0.041 \pm 0.007$ & $0.042 \pm 0.008$\\ 
        \quad GA Structural distance      & $0.041 \pm 0.024$ & $0.042 \pm 0.027$\\ 
        \quad NSGA-II Hybrid distance     & \fbox{$0.040 \pm 0.003$} & \fbox{$0.041 \pm 0.004$}\\
        \quad NSGA-II Structural distance & $0.069 \pm 0.050$ & $0.072 \pm 0.056$\\
        \quad NSGA-II Tree Complexity     & $0.110 \pm 0.052$ & $0.116 \pm 0.055$\\
        \quad NSGA-II Tree length         & $0.098 \pm 0.082$ & $0.106 \pm 0.082$\\
        \quad NSGA-II Nested tree length  & $0.086 \pm 0.078$ & $0.090 \pm 0.080$\\
        \textbf{Poly-10} & &\\
        \quad GA Standard                 & $0.173 \pm 0.282$ & $0.172 \pm 0.373$\\
        \quad GA Hybrid distance          & \fbox{$0.125 \pm 0.086$} & \fbox{$0.146 \pm 0.102$}\\ 
        \quad GA Structural distance      & $0.171 \pm 0.031$ & $0.186 \pm 0.073$\\ 
        \quad NSGA-II Hybrid distance     & \fbox{$0.128 \pm 0.070$} & \fbox{$0.147 \pm 0.107$}\\
        \quad NSGA-II Structural distance & $0.177 \pm 0.073$ & $0.195 \pm 0.109$\\
        \quad NSGA-II Tree Complexity     & $0.183 \pm 0.304$ & $0.187 \pm 0.329$\\
        \quad NSGA-II Tree length         & $0.187 \pm 0.234$ & $0.209 \pm 0.379$\\
        \quad NSGA-II Nested tree length  & $0.330 \pm 0.252$ & $0.383 \pm 0.412$\\
        \textbf{Pagie-1} & &\\
        \quad GA Standard                 & $0.003 \pm 0.004$ & $0.074 \pm 0.160$\\
        \quad GA Hybrid distance          & $0.003 \pm 0.003$ & $0.010 \pm 0.117$\\ 
        \quad GA Structural distance      & $0.004 \pm 0.002$ & \fbox{$0.005 \pm 0.005$}\\ 
        \quad NSGA-II Hybrid distance     & \fbox{$0.001 \pm 0.001$} & $0.007 \pm 0.017$\\
        \quad NSGA-II Structural distance & $0.003 \pm 0.004$ & \fbox{$0.005 \pm 0.027$}\\
        \quad NSGA-II Tree Complexity     & $0.007 \pm 0.006$ & $0.007 \pm 0.004$\\
        \quad NSGA-II Tree length         & $0.008 \pm 0.017$ & $0.009 \pm 0.028$\\
        \quad NSGA-II Nested tree length  & $0.006 \pm 0.009$ & $0.007 \pm 0.035$\\
        \bottomrule
    \end{tabular}
\end{table}

\begin{table}
    \centering
    \caption{Vladislavleva problems - median nmse $\pm$ iqr}\label{tab:experimental-results-vladislavleva}
    \begin{tabular}{lcc}
        & \textbf{Training} & \textbf{Test}\\
        \toprule
        \textbf{Vladislavleva-1} & &\\
        \quad GA Standard                 & $0.001 \pm 0.002$ & $0.046 \pm 0.179$\\
        \quad GA Hybrid distance          & $0.001 \pm 0.001$ & \fbox{$0.011 \pm 0.017$}\\ 
        \quad GA Structural distance      & $0.002 \pm 0.002$ & $0.028 \pm 0.031$\\ 
        \quad NSGA-II Hybrid distance     & \fbox{$0.000 \pm 0.000$} & $0.015 \pm 0.027$\\
        \quad NSGA-II Structural distance & $0.002 \pm 0.003$ & $0.045 \pm 0.056$\\
        \quad NSGA-II Tree Complexity     & $0.002 \pm 0.002$ & $0.032 \pm 0.087$\\
        \quad NSGA-II Tree length         & $0.001 \pm 0.002$ & $0.017 \pm 0.025$\\
        \quad NSGA-II Nested tree length  & $0.001 \pm 0.002$ & $0.017 \pm 0.017$\\
        \textbf{Vladislavleva-2} & &\\
        \quad GA Standard                 & $0.012 \pm 0.021$ & $0.015 \pm 0.047$\\
        \quad GA Hybrid distance          & $0.004 \pm 0.013$ & $0.009 \pm 0.019$\\ 
        \quad GA Structural distance      & $0.005 \pm 0.017$ & $0.010 \pm 0.025$\\ 
        \quad NSGA-II Hybrid distance     & \fbox{$0.001 \pm 0.002$} & \fbox{$0.002 \pm 0.004$}\\
        \quad NSGA-II Structural distance & $0.008 \pm 0.028$ & $0.013 \pm 0.032$\\
        \quad NSGA-II Tree Complexity     & $0.011 \pm 0.022$ & $0.021 \pm 0.041$\\
        \quad NSGA-II Tree length         & $0.004 \pm 0.009$ & $0.008 \pm 0.013$\\
        \quad NSGA-II Nested tree length  & $0.004 \pm 0.007$ & $0.006 \pm 0.011$\\
        \textbf{Vladislavleva-3} & &\\
        \quad GA Standard                 & $0.022 \pm 0.063$ & $0.063 \pm 0.113$\\
        \quad GA Hybrid distance          & $0.017 \pm 0.025$ & $0.018 \pm 0.054$\\ 
        \quad GA Structural distance      & $0.049 \pm 0.075$ & $0.076 \pm 0.385$\\ 
        \quad NSGA-II Hybrid distance     & \fbox{$0.004 \pm 0.011$} & \fbox{$0.008 \pm 0.064$}\\
        \quad NSGA-II Structural distance & $0.027 \pm 0.057$ & $0.111 \pm 0.245$\\
        \quad NSGA-II Tree Complexity     & $0.018 \pm 0.021$ & $0.022 \pm 0.050$\\
        \quad NSGA-II Tree length         & $0.012 \pm 0.015$ & $0.017 \pm 0.018$\\
        \quad NSGA-II Nested tree length  & $0.013 \pm 0.012$ & $0.012 \pm 0.011$\\
        \textbf{Vladislavleva-4} & &\\
        \quad GA Standard                 & $0.045 \pm 0.026$ & $0.096 \pm 0.047$\\
        \quad GA Hybrid distance          & $0.043 \pm 0.030$ & $0.095 \pm 0.053$\\ 
        \quad GA Structural distance      & $0.043 \pm 0.046$ & $0.108 \pm 0.091$\\ 
        \quad NSGA-II Hybrid distance     & $0.030 \pm 0.042$ & $0.060 \pm 0.068$\\
        \quad NSGA-II Structural distance & $0.032 \pm 0.048$ & $0.096 \pm 0.083$\\
        \quad NSGA-II Tree Complexity     & $0.046 \pm 0.028$ & $0.093 \pm 0.056$\\
        \quad NSGA-II Tree length         & \fbox{$0.024 \pm 0.060$} & \fbox{$0.051 \pm 0.111$}\\
        \quad NSGA-II Nested tree length  & $0.030 \pm 0.036$ & $0.067 \pm 0.083$\\
        \textbf{Vladislavleva-5} & &\\
        \quad GA Standard                 & $0.001 \pm 0.003$ & $0.013 \pm 0.132$\\
        \quad GA Hybrid distance          & $0.001 \pm 0.002$ & $0.004 \pm 0.018$\\ 
        \quad GA Structural distance      & $0.003 \pm 0.004$ & $0.015 \pm 0.148$\\ 
        \quad NSGA-II Hybrid distance     & \fbox{$0.000 \pm 0.001$} & \fbox{$0.002 \pm 0.010$}\\
        \quad NSGA-II Structural distance & $0.003 \pm 0.007$ & $0.140 \pm 0.172$\\
        \quad NSGA-II Tree Complexity     & $0.022 \pm 0.170$ & $0.109 \pm 0.200$\\
        \quad NSGA-II Tree length         & $0.002 \pm 0.010$ & $0.009 \pm 0.026$\\
        \quad NSGA-II Nested tree length  & $0.002 \pm 0.004$ & $0.009 \pm 0.013$\\
        \textbf{Vladislavleva-6} & &\\
        \quad GA Standard                 & $0.112 \pm 0.220$ & $1.372 \pm 4.547$\\
        \quad GA Hybrid distance          & $0.057 \pm 0.140$ & $0.486 \pm 1.731$\\ 
        \quad GA Structural distance      & $0.068 \pm 0.160$ & $0.543 \pm 1.795$\\ 
        \quad NSGA-II Hybrid distance     & \fbox{$0.000 \pm 0.000$} & \fbox{$0.000 \pm 0.000$}\\
        \quad NSGA-II Structural distance & $0.124 \pm 0.167$ & $0.953 \pm 1.572$\\
        \quad NSGA-II Tree Complexity     & $0.036 \pm 0.068$ & $1.445 \pm 6.467$\\
        \quad NSGA-II Tree length         & \fbox{$0.000 \pm 0.039$} & \fbox{$0.000 \pm 0.396$}\\
        \quad NSGA-II Nested tree length  & \fbox{$0.000 \pm 0.088$} & \fbox{$0.000 \pm 0.796$}\\
        \textbf{Vladislavleva-7} & &\\
        \quad GA Standard                 & $0.099 \pm 0.032$ & $0.118 \pm 0.068$\\
        \quad GA Hybrid distance          & $0.092 \pm 0.017$ & \fbox{$0.104 \pm 0.020$}\\ 
        \quad GA Structural distance      & $0.105 \pm 0.022$ & $0.126 \pm 0.070$\\ 
        \quad NSGA-II Hybrid distance     & \fbox{$0.079 \pm 0.011$} & $0.108 \pm 0.071$\\
        \quad NSGA-II Structural distance & $0.103 \pm 0.022$ & $0.147 \pm 0.074$\\
        \quad NSGA-II Tree Complexity     & $0.097 \pm 0.085$ & $0.107 \pm 0.034$\\
        \quad NSGA-II Tree length         & $0.094 \pm 0.036$ & $0.112 \pm 0.047$\\
        \quad NSGA-II Nested tree length  & $0.093 \pm 0.024$ & $0.106 \pm 0.034$\\
        \textbf{Vladislavleva-8} & &\\
        \quad GA Standard                 & $0.181 \pm 0.223$ & $0.642 \pm 0.402$\\
        \quad GA Hybrid distance          & $0.033 \pm 0.235$ & $0.520 \pm 0.440$\\ 
        \quad GA Structural distance      & $0.020 \pm 0.233$ & $0.480 \pm 0.399$\\ 
        \quad NSGA-II Hybrid distance     & \fbox{$0.007 \pm 0.019$} & \fbox{$0.427 \pm 0.608$}\\
        \quad NSGA-II Structural distance & $0.039 \pm 0.100$ & $0.488 \pm 0.419$\\
        \quad NSGA-II Tree Complexity     & $0.013 \pm 0.020$ & $0.759 \pm 0.638$\\
        \quad NSGA-II Tree length         & $0.030 \pm 0.038$ & $0.744 \pm 0.617$\\
        \quad NSGA-II Nested tree length  & $0.011 \pm 0.017$ & $0.534 \pm 0.559$\\
        \bottomrule
    \end{tabular}
\end{table}

Tables~\ref{tab:experimental-results-various} and~\ref{tab:experimental-results-vladislavleva} summarize the obtained results, highlighting the best training and test result for each problem.
Except for the tree distance, all other NSGA-II secondary objectives were designed with the purpose of reducing model complexity and are therefore expected to return simpler, if slightly worse solutions.

The \emph{NSGA-II Hybrid distance} configuration produces the best training result on 12 out of 13 problem instances and the best test results on 9 out of 13 instances. Our proposed diversity criterion using the hybrid tree distance is able to improve solution quality on the majority of tested problems, while at the same time providing more robustness (in terms of the reported IQR which indicates lower dispersion of the results). 

Somewhat surprisingly, \emph{GA Hybrid Distance} provides the second-best performance, placing just after \emph{NSGA-II Hybrid distance} when considering the median rank over all tested problems, as shown in Table~\ref{tab:average-rank}. 

\emph{NSGA-II Structural distance} and \emph{GA Structural distance} employing the structural distance metric are not up to par with their their hybrid-semantic counterparts, suggesting that purely structural diversity does not seem to guide the search in the right direction, although \emph{GA Structural distance} manages to outperform the standard GA algorithm. \emph{NSGA-II Structural distance} places last in our ranking, indicating that the structural tree distance does not work well together with the crowding distance already employed by the NSGA algorithm. 

\begin{table}
    \centering
    \caption{Median algorithm rank over all problems.}\label{tab:average-rank}
    \begin{tabular}{lcc}
        \toprule
        Algorithm & Training rank & Test rank\\
        \midrule
        NSGA-II Hybrid distance          & $1.0$ & $1.0$\\
        GA Hybrid distance               & $2.0$ & $3.0$\\
        NSGA-II Nested tree size         & $4.0$ & $4.0$\\
        GA Structural distance           & $5.0$ & $5.0$\\
        NSGA-II Tree size                & $5.0$ & $5.0$\\
        GA Standard                      & $6.0$ & $6.0$\\
        NSGA-II Tree complexity          & $6.0$ & $6.0$\\
        NSGA-II Structural distance      & $7.0$ & $6.0$\\
        \bottomrule
    \end{tabular}
\end{table}


Overall, the results validate the hybrid diversity metric as a viable approach for fine-tuning the algorithm's exploration capabilities. We hypothesize that recombination operators are more effective in producing better solutions by combining relevant traits from more diverse parents, when diversity is defined as a combination of structure and semantics. We further investigate this aspect by calculating the evolution of population diversity in each generation for a selected problem.


\subsection{Evolution of Diversity}
We focus on one of the benchmark problems without noise, the Poly-10, and calculate average similarity (as defined in Equation~\ref{eqn:tree-similarity}) in each generation. We express the evolution of diversity in terms of average similarity and discuss it's relationship with average tree length. 

\begin{figure}[ht]
    \caption{\sc{Structural similarity}}\label{fig:structural-similarity}
    \includegraphics[width=\linewidth]{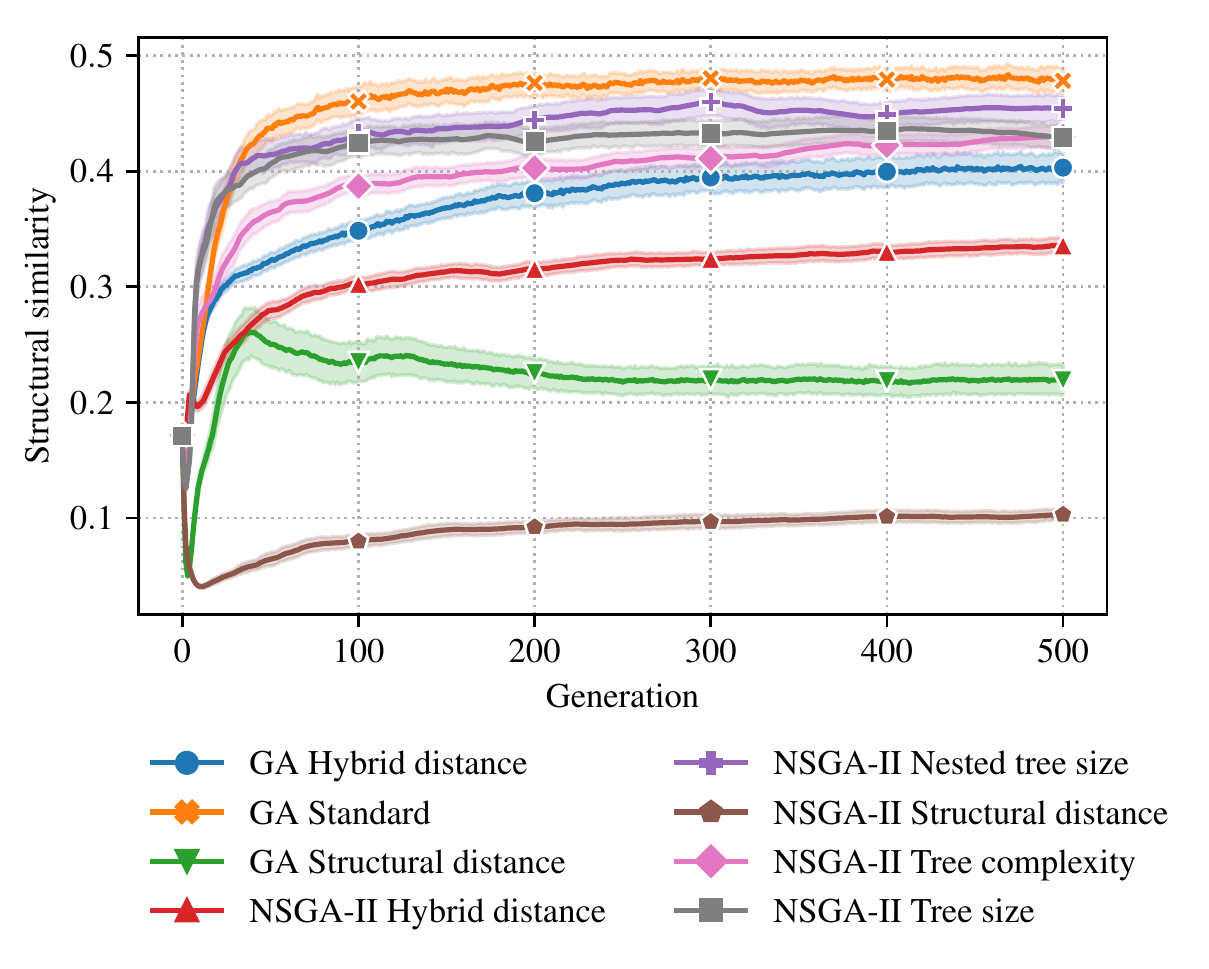}
\end{figure}

\begin{figure}[ht]
    \caption{\sc{Hybrid similarity}}\label{fig:hybrid-similarity}
    \includegraphics[width=\linewidth]{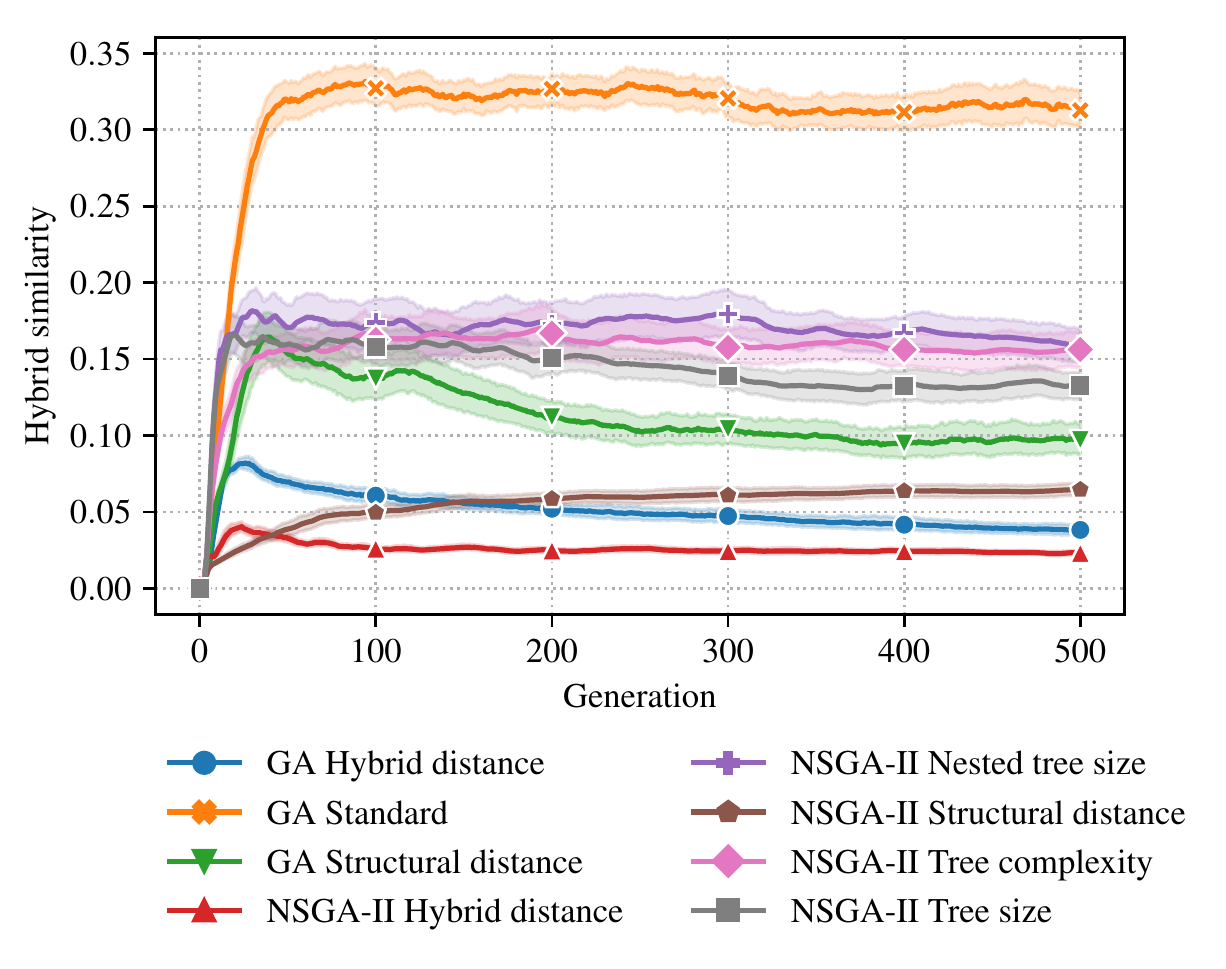}
\end{figure}

Subsection~\ref{subsec:benchmark-results} has already shown that the hybrid tree distance is a more effective objective than the structural tree distance. Figures~\ref{fig:structural-similarity} and~\ref{fig:hybrid-similarity} show the evolution of the two measures over the generations, showing the average value per generation and the $95\%$ confidence region around the average.

\begin{figure}[ht]
    \caption{\sc{Average tree length}}\label{fig:tree-length}
    \includegraphics[width=\linewidth]{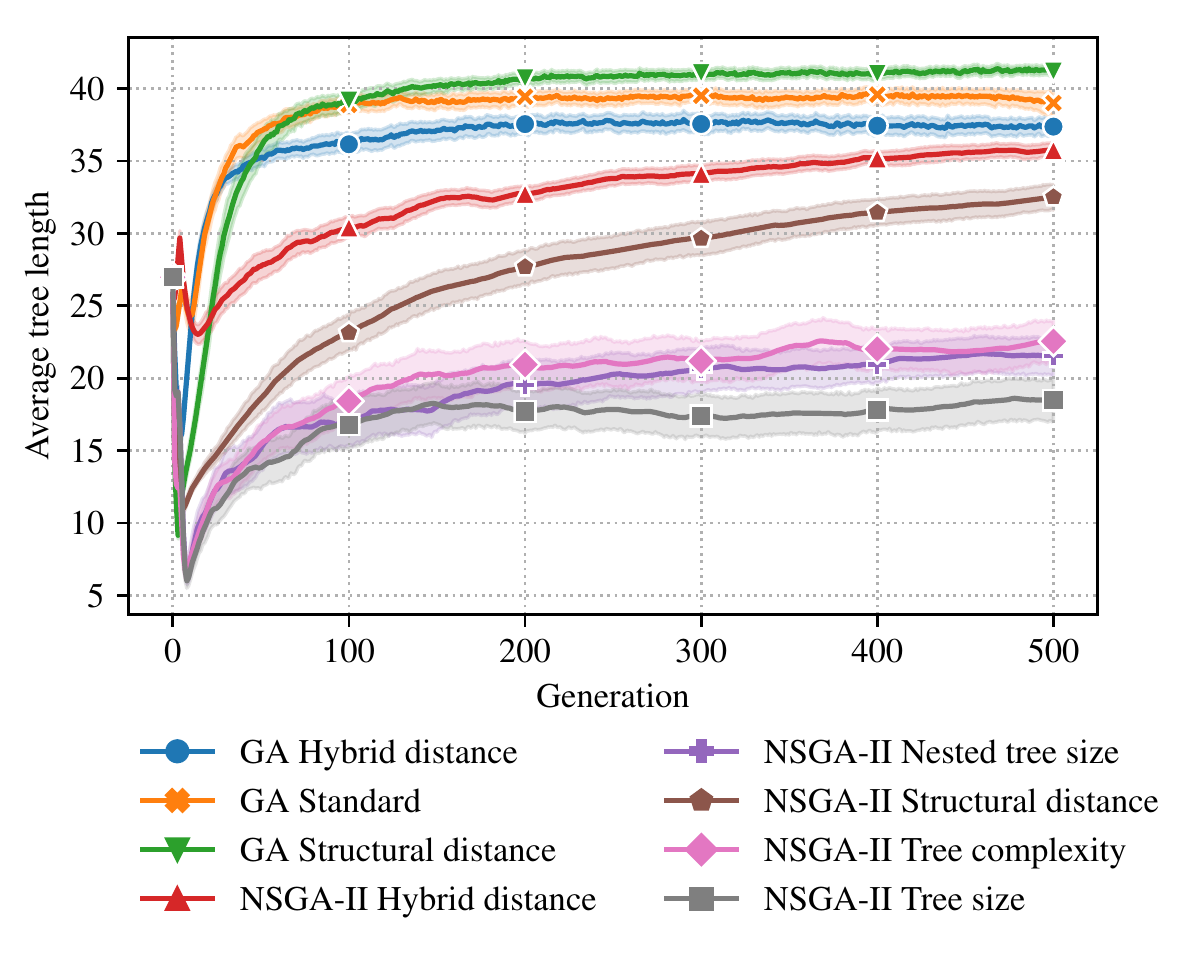}
\end{figure}

As expected, explicit selection for structural diversity directly leads to low structural similarity. However, when comparing \emph{GA Structural distance} with \emph{NSGA-II Structural distance} we notice that the former exhibits significantly higher structural similarity than the latter. This can be explained by the different ways in which this selection pressure is applied. In the standard GA, quality improvements in the beginning of the run are large enough to outweigh the penalty incurred by the diversity term, leading to an increase in structural similarity. 

The NSGA-II on the other hand keeps non-dominated solutions with lower quality but high diversity, thus leading to a decrease in structural similarity.
We observe the same behavior on the hybrid similarity curves in Figure~\ref{fig:hybrid-similarity} where the \emph{NSGA-II Hybrid distance} displays overall lower similarity levels.

The two graphs illustrate an interesting relationship between the structural and hybrid distance measures: configurations explicitly selecting for structural diversity (\emph{GA Structural distance} and \emph{NSGA-II Structural distance}) display low structural similarity but increased hybrid similarity. Conversely, configurations explicitly selecting for hybrid diversity (\emph{GA Hybrid distance} and \emph{NSGA-II Hybrid distance}) display low hybrid similarity but increased structural similarity.

\begin{figure*}
    \caption{\sc{Tree similarity heatmap in the last generation, averaged over 50 runs for each configuration. Lower triangular matrix represents structural pairwise similarities, while upper triangular matrix represents hybrid pairwise similarities.}}\label{fig:similarity-heatmap}
    \includegraphics[width=\textwidth]{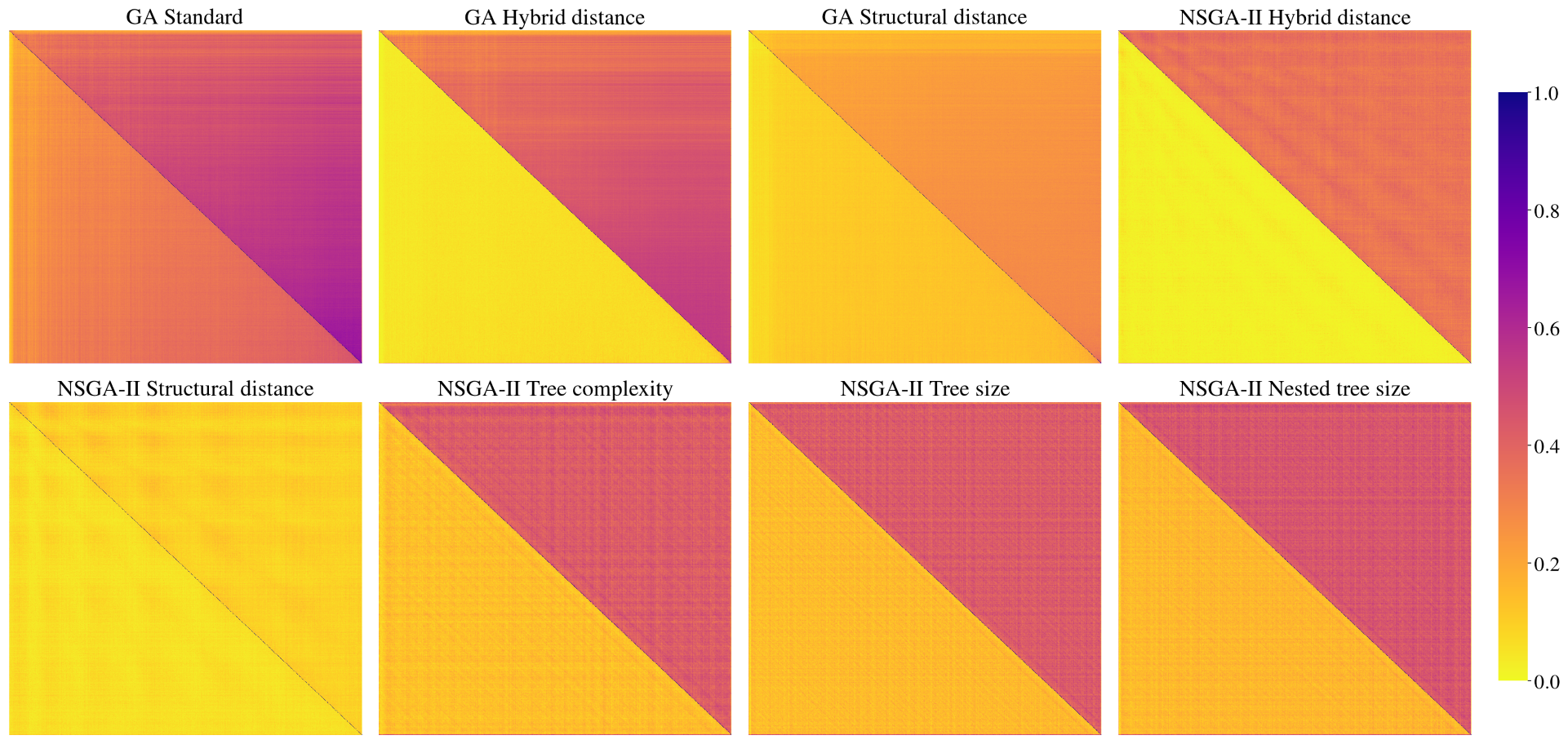}
\end{figure*}

This indicates that structural diversity does not imply semantic diversity, and vice versa; and confirms that a hybrid measure as suggested by~\cite{Burks2018} represents a more effective approach for the pursuit of diversity. This slightly counter-intuitive relationship is illustrated in Figure~\ref{fig:similarity-heatmap} where the two distance measures are graphically compared by sharing the upper and lower triangular halves of the same heatmap.

Finally, Figure~\ref{fig:tree-length} shows that \emph{GA} tends to produce larger trees compared to \emph{NSGA}. However, the relationship between the average tree length and the proposed diversity measures differs between the two algorithms:
\begin{itemize}
    \item In the case of \emph{GA}, structural diversity promotes larger trees while hybrid diversity promotes smaller trees compared to the standard configuration.
    \item In the case of \emph{NSGA}, structural diversity promotes smaller trees while hybrid diversity promotes larger trees, both configurations remaining overall under the level of \emph{GA} average tree length. The remaining three configurations (\emph{tree complexity, tree size} and \emph{nested tree size}) explicitly select for low size, thus are not directly comparable.
\end{itemize}

The parsimony-focused NSGA variants (using tree complexity, tree size and nested tree size) all display lower average tree length, but at the same time they display higher average similarity (structural and hybrid) than the tree distance-based configurations.

Overall, we do not find a clear correlation between tree length and similarity although this aspect deserves further consideration.

The heatmaps in Figure~\ref{fig:similarity-heatmap} are constructed from the population similarity matrix in the last generation of the algorithm, averaged over 50 repetitions. The rows and columns of the matrix are ordered according to the increasing fitness values of the individuals. The heatmaps for \emph{GA Standard}, \emph{GA Hybrid distance} and, to a lesser degree, \emph{GA Structural distance} suggest a linear correlation between diversity and fitness, with more fit individuals having lower average distance to the rest of the population (visible in the graphs as a left-to-right or top-to-bottom gradient). This relationship is not apparent on any of the \emph{NSGA} configurations, where Pareto-front domination dynamics lead to a more uniform distribution of similarity and fitness values.

\subsection{Runtime Overhead}
We estimate runtime overhead by comparing wall-clock time required for the computation of the tree distance matrix with the total runtime of the genetic algorithm. We use two differently-sized benchmark problems: the \emph{Poly-10} ($250$ training samples) and \emph{Breiman-I} ($5000$ training samples), in order to determine how this overhead relates to problem dimensions. We report median values over 50 repetitions in Table~\ref{tab:execution-time}, where the last column shows the difference to baseline (GA Standard), discounting the overhead.

We additionally profile the computation of the distance matrix and identify the critical code path. 
We run an independent test using $N=1000$ randomly initialized trees with of length $50$.
Our results show that the computation of the co-occurrence count shown in Algorithm~\ref{alg:tree-distance} dominates the runtime of the tree distance algorithm (accounting for more than $90\%$ of the computation time). These benchmark results are shown in Table~\ref{tab:tree-distance-benchmark}.

Tables~\ref{tab:execution-time} and~\ref{tab:tree-distance-benchmark} reveal a couple of interesting phenomena. On the one hand, the computation of the hybrid tree distance is slightly more expensive than the structural tree distance, due to the different characteristics of the input hash value sequences which influence the of runtime of Algorithm~\ref{alg:tree-distance} (structural hashing leads to a higher probability of repeated values due to leaf nodes being hashed based only on their label, as opposed to hybrid hashing where leaf node coefficients are also hashed, leading to more diverse hash sequences).

On the other hand, different algorithm dynamics and average tree lengths cause the structural GA and NSGA variants to take more time than their hybrid counterparts. Overall, the overhead determined by the computation of the tree distance matrix does not exceed $200$ms per generation, making this approach feasible for online diversity steering. Compared to the standard GA, the difference becomes relatively big only when the problem size is very small ($250$ rows, Poly-10 problem), in which case runtime is usually not a concern. For the larger problem ($5000$ rows, Breiman-I problem) the relative overhead ranges from $15\%$ (GA Hybrid distance) to $30\%$ (GA Structural distance). 

Subtracting the distance computation overhead from the total running time (last column of Table~\ref{tab:execution-time}) further shows that the increase in total runtime is caused not only by the computation of the distance matrix, but also by changes in algorithm dynamics.  

\begin{table}[ht]
    \caption{Algorithm median execution time, distance computation overhead, and difference to baseline (in seconds).}\label{tab:execution-time}
    \centering
    \begin{tabular}{lrrr}
        \toprule
        Algorithm & Runtime (s) & Overhead (s) & Diff. (s)\\
        \midrule
        \textbf{Poly-10} & & &\\
        GA Standard (baseline)      & $160.1$ &                  \\
        GA Hybrid distance          & $285.1$ & $134.6$ & $-9.6$ \\
        GA Structural distance      & $301.9$ & $143.3$ & $-1.5$ \\
        NSGA-II Hybrid distance     & $317.5$ & $109.5$ & $48.0$ \\
        NSGA-II Structural distance & $293.1$ & $87.3$  & $45.7$ \\
        NSGA-II Complexity          & $195.5$ &         &        \\
        NSGA-II Nested Tree Length  & $196.5$ &         &        \\
        NSGA-II Tree Length         & $189.8$ &         &        \\
        \midrule
        \textbf{Breiman-I} & &\\
        GA Standard (baseline)      & $899.3$  &                   \\
        GA Hybrid distance          & $1034.9$ & $138.1$ & $-2.5$  \\
        GA Structural distance      & $1204.1$ & $142.9$ & $161.8$ \\
        NSGA-II Hybrid distance     & $1092.0$ & $109.2$ & $83.4$  \\
        NSGA-II Structural distance & $1140.6$ & $91.9 $ & $149.4$ \\
        NSGA-II Complexity          & $830.4$  &         &         \\
        NSGA-II Nested Tree Length  & $842.0$  &         &         \\
        NSGA-II Tree Length         & $820.3$  &         &         \\
        \bottomrule
    \end{tabular}
\end{table}

\begin{table}[ht]
    \caption{Runtime of structural and hybrid distance matrix computation (in milliseconds)}\label{tab:tree-distance-benchmark}
    \centering
    \begin{tabular}{lrr}
        \toprule
        Procedure & Structural & Hybrid\\
        \midrule
        $-$ Compute hash value sequences & $10.11$ & $10.61$\\
        \quad $\circ$ Sort child subarrays & $8.96$ & $9.27$\\
        \quad $\circ$ Sort final hash value sequence & $1.15$ & $1.34$\\
        $-$ Compute distance (co-occurrence count) & $126.50$ & $156.40$\\[0.2ex]
        \textbf{Total} & $136.61$ & $167.01$\\
        \bottomrule
    \end{tabular}
\end{table}

\section{Conclusion}\label{sec:conclusion}
In this paper, we introduced an efficient method for the computation of tree distances and demonstrated its usefulness as an online strategy for diversity control. 

The proposed method employs tree hashing in order to convert tree individuals into sequences of hash values, making it particularly suited for en-masse computation of pairwise tree distances. In this particular usage scenario, we were able to achieve a two orders of magnitude speed improvement over a similar method from the literature.

We compared two kinds of distance supported by our implementation, namely a purely structural distance and a hybrid distance taking into account both structure and numeric parameters.
We investigated the hypothesis often-encountered in the literature that behavioural diversity plays a more important role in achieving good performance than structural diversity.

Empirical results on a suite of symbolic regression benchmark problems support the hypothesis that, on the one hand, diversity-steering provides a real benefit in terms of solution quality compared to the standard GA and the parsimony-oriented NSGA variants. On the other hand, our results show that a hybrid approach (attaching semantics to an inherently structural diversity metric) is clearly superior to the simple structural-genotypic approach. 

We further analyzed the runtime impact of our proposed approach and showed that a) distance matrix computation overhead is not significant, and b) subtle changes in algorithm dynamics also have an influence on execution time. 

Due to its simplicity, our method can be further integrated with other genetic algorithm variants, either by adding it as a penalty to the fitness function or by integrating the concept of crowding distance (eg., during selection). 

\bibliographystyle{IEEEtran}
\bibliography{bibliography}

\end{document}